\documentclass[10pt,twocolumn,letterpaper]{article}

\usepackage{iccv}
\usepackage{times}
\usepackage{epsfig}
\usepackage{graphicx}
\usepackage{amsmath}
\usepackage{amssymb}

\usepackage{comment}
\usepackage{multirow}
\usepackage[dvipsnames]{xcolor} % Remove before submission if necessary
\usepackage{capt-of}
\usepackage{stfloats}
\usepackage[pagebackref=true,breaklinks=true,letterpaper=true,colorlinks,bookmarks=false]{hyperref}
\iccvfinalcopy % *** Uncomment this line for the final submission

\ificcvfinal\fi

\begin{document}

%%%%%%%%% TITLE
\title{PoseRAC: Pose Saliency Transformer for Repetitive Action Counting}

\author{Ziyu Yao, Xuxin Cheng, Yuexian Zou\\
School of Electronic and Computer Engineering, Peking University, China\\
\{yaozy, chengxx\}@stu.pku.edu.cn; zouyx@pku.edu.cn\\
}

\let\oldtwocolumn\twocolumn
\renewcommand\twocolumn[1][]{%
    \oldtwocolumn[{#1}{
        \begin{center}
            \vspace{-13pt}
            \includegraphics[trim=0.0cm 0.0cm 0cm 0.0cm,width=0.98\textwidth]{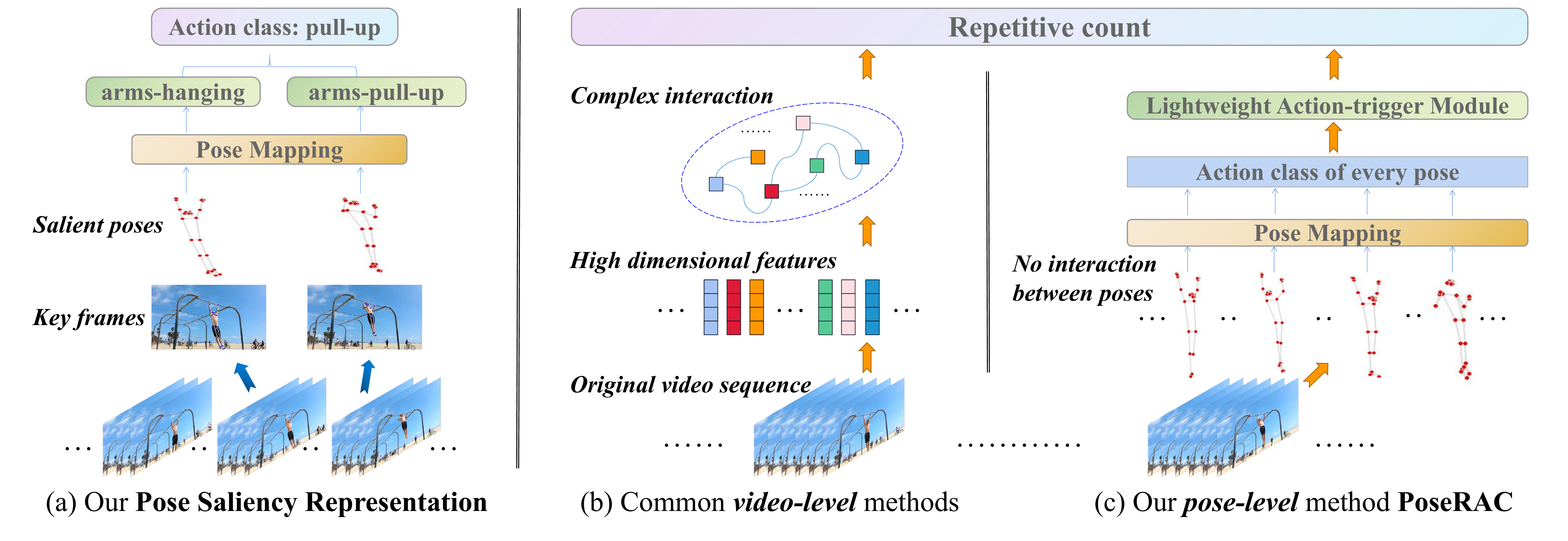}
            \end{center}
            \vspace{-13pt}
            \captionof{figure}{
            \textbf{(a)} Pose Saliency Representation is our proposed mechanism using the two most salient poses to represent each action, which captures essential information while reducing computational overhead. \textbf{(b)} Video-level methods require interacting intra-frame and inter-frame information, leading to increasing calculation and potentially computing useless information. \textbf{(c)} PoseRAC is our proposed pose-level method which estimates the pose of each frame and utilizes such core information to classify actions frame-by-frame. Unlike video-level methods, our approach reduces calculation and uses a lightweight action-trigger to obtain the final result.
            }
            \label{fig1}
            \vspace{10pt}
    }]
}

\maketitle
\ificcvfinal\thispagestyle{empty}\fi

%%%%%%%%% ABSTRACT
\begin{abstract}

This paper presents a significant contribution to the field of repetitive action counting through the introduction of a new approach called \textbf{Pose Saliency Representation}. The proposed method efficiently represents each action using only two salient poses instead of redundant frames, which significantly reduces the computational cost while improving the performance. Moreover, we introduce a pose-level method, \textbf{PoseRAC}, which is based on this representation and achieves state-of-the-art performance on two new version datasets by using \textbf{Pose Saliency Annotation} to annotate salient poses for training. Our lightweight model is highly efficient, requiring only 20 minutes for training on a GPU, and infers nearly 10x faster compared to previous methods. In addition, our approach achieves a substantial improvement over the previous state-of-the-art TransRAC, achieving an OBO metric of 0.56 compared to 0.29 of TransRAC. The code and new dataset are available at \href{https://github.com/MiracleDance/PoseRAC}{https://github.com/MiracleDance/PoseRAC} for further research and experimentation, making our proposed approach highly accessible to the research community.

\end{abstract}
\section{Introduction}

% \begin{figure*}[t]]
% \centering
% \includegraphics[width=1.0\textwidth]{figure1.pdf}
% \caption{\textbf{(a)} Pose Saliency Representation is our proposed mechanism using the two most salient poses to represent an action, which captures essential information while reducing computational overhead. \textbf{(b)} Video-level methods require interacting intra-frame and inter-frame information, leading to increasing calculation and potentially computing useless information. \textbf{(c)} PoseRAC is our proposed pose-level method which estimates the pose of each frame and utilizes such core information to classify actions frame-by-frame. Unlike video-level methods, our approach reduces calculation and uses a lightweight action-trigger to obtain the final result.}
% \label{fig1}
% \end{figure*}

Periodic movement is a ubiquitous phenomenon in nature, including human activities. Repetitive action counting aims to count the number of repetitive actions in a video, which is crucial for analyzing human action, such as pedestrian detection\cite{ran2007pedestrian}, camera calibration\cite{huang2016camera}, and three-dimensional reconstruction\cite{ribnick20103d, li2018structure}. However, this task has not been extensively explored. Previous works, such as those by Levy et al. \cite{levy2015live} and Pogalin et al. \cite{pogalin2008visual}, rely on the periodicity assumption. Still, the period of an action can be long or short or even mixed with inconsistent cycles. Other works, such as those by Zhang et al. \cite{zhang2020context} and Dwibedi et al. \cite{dwibedi2020counting}, exploit contextual information to achieve better results. Moreover, Hu et al. \cite{hu2022transrac} propose a multi-scale temporal correlation encoder that achieves state-of-the-art performance, maintaining it until our work.

All current works on this task are video-level, which involves expensive feature extraction and sophisticated video-context interaction. Meanwhile, the research on human pose estimation is in full swing, and the current results, such as those by Bazarevsky et al. \cite{bazarevsky2020blazepose} and Xu et al. \cite{xu2022vitpose}, can already support practical applications. However, human pose has not been well-explored in repetitive action counting tasks. We believe that if human poses can be well-utilized, it can greatly improve the performance and efficiency of this task.

\begin{figure}[t]
\centering
\includegraphics[width=0.49\textwidth]{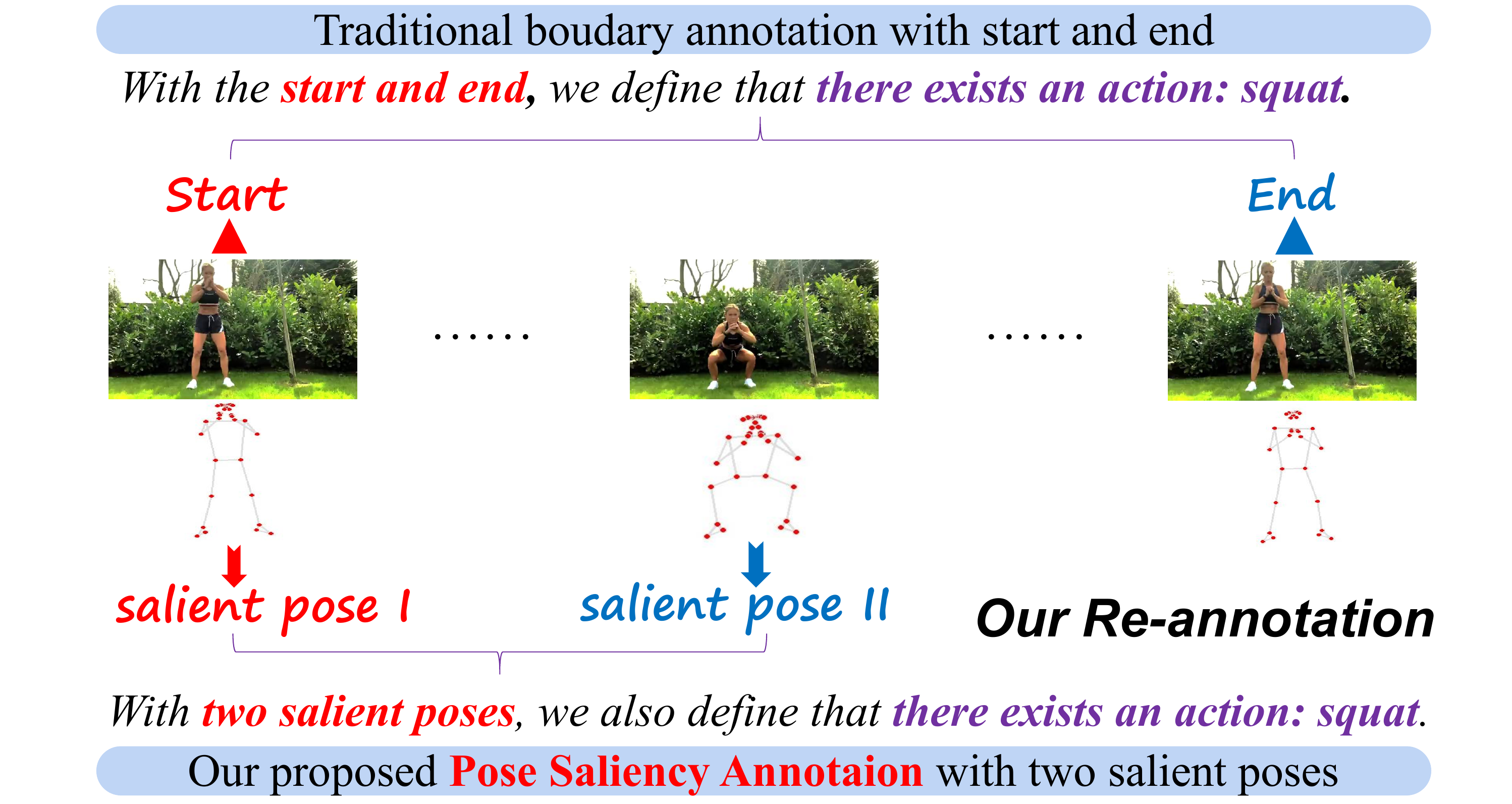}
\caption{Our Pose Saliency Annotation. Instead of annotating the start and end, we annotate the two most salient frames.}
\label{fig2}
\end{figure}

Our motivations arise from two aspects:

\begin{itemize}

\item {\bf How to be more effective?} Human body pose is the most essential factor in an action, and other information sometimes interferes with the model, such as hue, face, skin color, etc. If we ignore those factors and focus on the most essential poses, we can improve the performance on this work. Such idea of using saliency information can be found in other tasks\cite{tian2022bi,ji2022dmra,cheng2023ssvmr}. 

\item {\bf How to be more efficient?} Video-level methods require complex contextual interactions for the entire video, including those irrelevant background features, which bear a huge amount of calculation. However, if we use dozens of human body key points to represent an action, it can greatly improve the speed of the model. The significant reduction in computing overhead can lead to greater application prospects.

\end{itemize}

Therefore, we propose a mechanism called {\bf P}ose {\bf S}aliency {\bf R}epresentation ({\bf PSR}), which represents an action using the two most salient poses. The common method of representing an action with RGB frames is redundant, as intra-frame spatial information and inter-frame temporal information need complex calculations to obtain high-level semantic information. However, our PSR mechanism, as shown in Figure \ref{fig1} (a), can use only two key frames to capture the salient pose features of each action and establish a unique mapping between salient poses and action classes.

Based on PSR, we propose the first pose-level network called {\bf Pose} Saliency Transformer for {\bf R}epetitive {\bf A}ction {\bf C}ounting ({\bf PoseRAC}), as shown in Figure \ref{fig1} (c). We first extract poses of all frames using state-of-the-art algorithms, such as BlazePose\cite{bazarevsky2020blazepose}. Next, our core model maps each of these poses to an action class, which we detail in Sec~$\S\ref{second}$ and~$\S\ref{third}$. Lastly, we design a lightweight Action-trigger module, where the action of each frame is used to complete video-level counting. Our PoseRAC does not require complicated calculations and has been experimentally proven to have the advantages of both accuracy and speed.

\begin{figure}[t]
\centering
\includegraphics[width=0.46\textwidth]{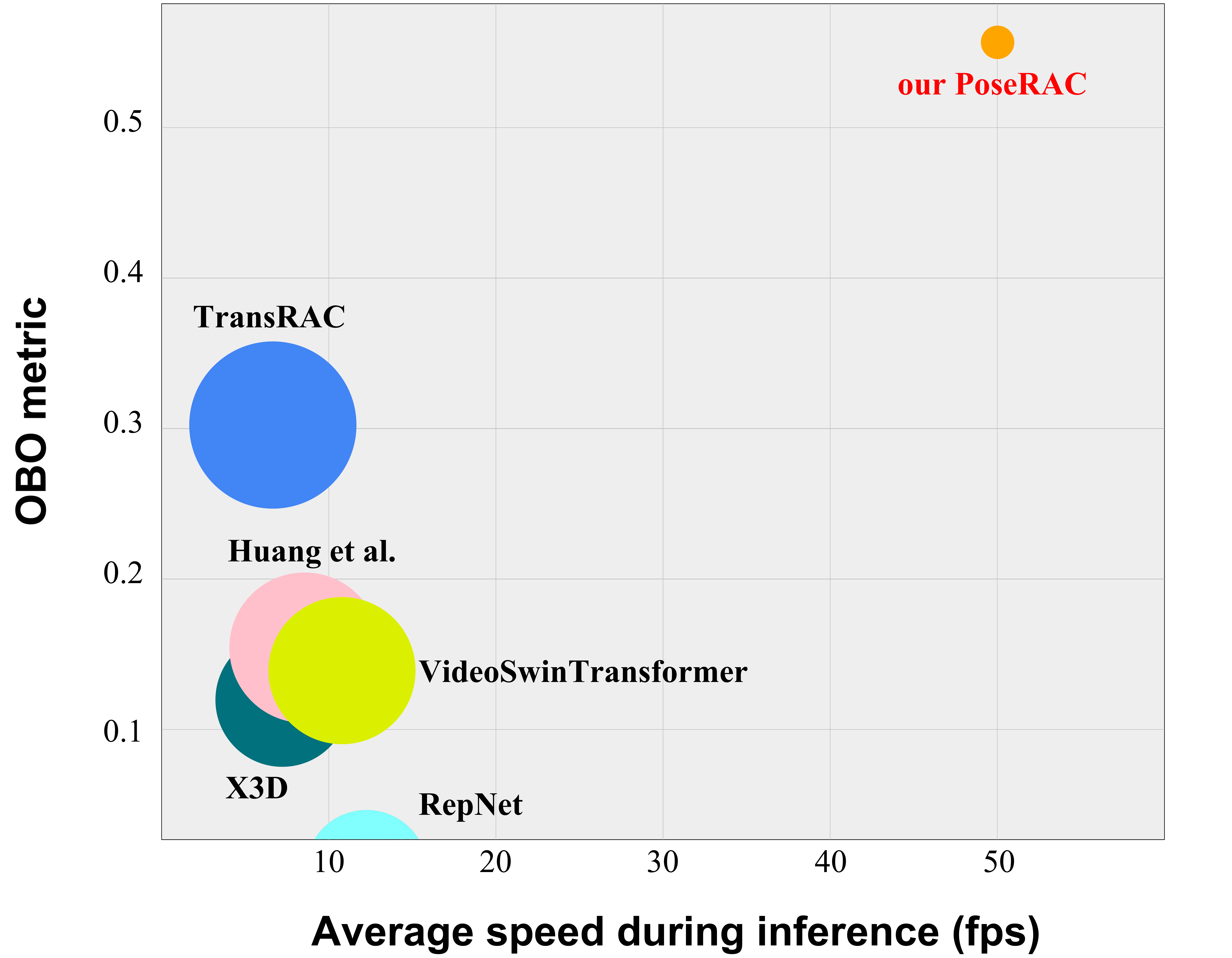}
\caption{The comparison of PoseRAC and previous SOTA methods on \emph{RepCount-(pose)} test set regarding speed, OBO accuracy and model size, which represented by the sizes of bubbles. When calculating the speed of PoseRAC, the time of pose estimation is included.}
\label{fig6}
\end{figure}

However, learning such mapping between salient poses and actions requires salient annotation, which current datasets, such as the \emph{RepCount}\cite{hu2022transrac}, do not provide. It only annotates the start and end frames of each action, which does not necessarily the most salient moments of an action. To solve this problem, we improve on the current datasets for pose-level method. Dataset augmentation has been proven to be effective\cite{zhang2018mixup, zhu2022dynamic, cheng2022m3st, li2023generating}. Different from the original annotation of \emph{RepCount}, we propose a new annotation idea that has not yet been explored: {\bf P}ose {\bf S}aliency {\bf A}nnotation ({\bf PSA}). As Figure \ref{fig2} shows, instead of annotating the start and end frames of each action in traditional boundary annotation, we annotate the two most salient frames. We first pre-define the two most salient poses for all action classes. For example, for sit-up, we define lying down and sitting as two salient poses, for squats, body-upright and squatting are two salient poses, and so on. Then for all videos in training set, we annotate the frame indices of two salient poses for each action event. With such novel annotation, we can obtain the most representative poses to greatly optimize our training process. Combining Pose Saliency Annotation, we augment two current datasets \emph{RepCount} and \emph{UCFRep} with pose-level annotations, and create two new version: {\bf \emph{RepCount-pose}} and {\bf \emph{UCFRep-pose}}, which can be used by all future pose-level methods.

\begin{figure*}[t]
\centering
\includegraphics[width=0.95\textwidth]{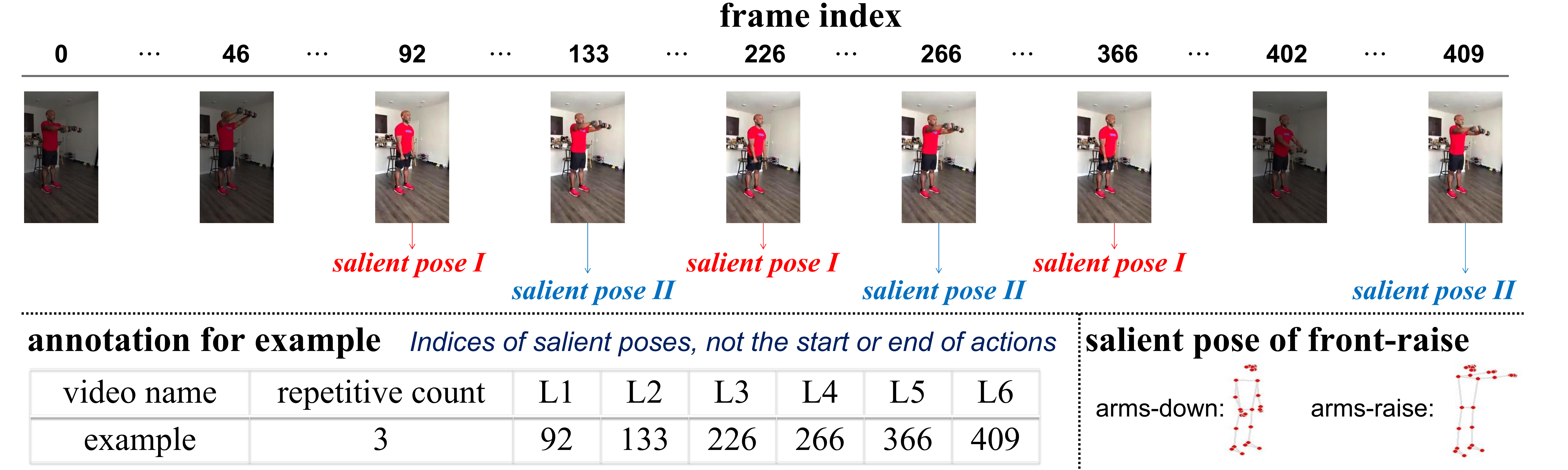}
\caption{{\bf The specific implementation of Pose Saliency Annotation.} We define two salient poses for each action and precisely select the frames in which these poses occur in a given video. Here we take front raise action as an example, and L1, L2 mean the indices of two salient poses of the first action, L3 to L6 are similar.}
\label{fig3}
\end{figure*}

During inference, we make a fair comparison with the previous approaches, and using PSA to train PoseRAC, we achieve new state-of-the-art performance on two test set, far outperforming all current methods, with an OBO metric of 0.56 compared to 0.29 of previous state-of-the-art TransRAC. Moreover, our model has a exaggerated running speed, which takes only 20 minutes to train on a single GPU, and it is even so lightweight to train in only one hour and a half on a CPU, which is unimaginable in previous video-level methods. PoseRAC is also very fast during inference, which is almost 10x faster than the previous state-of-the-art TransRAC on the average speed per frame. Figure \ref{fig6} further demonstrates the superiority of our method in terms of speed and accuracy, even having significantly fewer model parameters than previous video-level methods.

We summarize our contributions in three-fold:

\begin{itemize}

\item Based on our proposed Pose Saliency Annotation, we annotate the salient poses of all videos, augment two current datasets with pose-level annotations, and create two new version: \emph{RepCount-pose} and \emph{UCFRep-pose}.

\item We propose Pose Saliency Representation which represents each action with two salient poses, rather than redundant frames. Then we propose the first pose-level model PoseRAC to solve repetitive action counting task, which is simple, effective and efficient.

\item Our PoseRAC not only far outperforms all state-of-the-art methods in performance on current datasets, but also has particularly high efficiency. It is lightweight enough to be easily trained on CPU, which is undoubtedly not possible with previous methods.

\end{itemize}
\section{Related Works}

\noindent{\bf Repetitive action counting.} Early methods \cite{cutler2000robust, pogalin2008visual, tsai1994cyclic, albu2008generic, laptev2005periodic, lu2004repetitive, panagiotakis2018unsupervised, tralie2018quasi, chetverikov2006motion} focus on compressing the motion field into one-dimensional signals to recover the repetition, where Fourier analysis\cite{pogalin2008visual, tsai1994cyclic, briassouli2007extraction}, peak detection\cite{thangali2005periodic}, classification\cite{davis2000categorical, levy2015live} can be used. However, they are limited to stationary situations, so \cite{runia2019repetition, runia2018real} collect a dataset with non-stationary repetitions. As they all analyze the visual information, \cite{zhang2021repetitive} utilizes the sound for the first time. Moreover, \cite{zhang2020context} proposes a context-aware model and constructs \emph{UCFRep} dataset with 526 videos. Similarly, \cite{dwibedi2020counting} creates \emph{Countix} which contains over 6000 videos. However, they only have coarse-grained annotation, so \cite{hu2022transrac} introduces a large-scale \emph{RepCount} dataset with fine-grained annotation of the actions. This work also encodes multi-scale temporal correlation to improve the performance and efficiency.

Different from them, we propose a simple pose-level method which outperforms all previous work in the speed and performance. Meanwhile, as current datasets lack salient annotations, we re-annotate \emph{RepCount} and \emph{UCFRep} to contribute to the future pose-level work on this task.

\noindent{\bf Human pose estimation.} Convolutional nerual network is mainstream in early works\cite{xiao2018simple, rafi2016efficient, wei2016convolutional, sun2019deep}, but when vision transformer emerged in various visual tasks\cite{li2022exploring, liu2022video}, it began to be used more\cite{yang2021transpose, li2021tokenpose}. Here we focus on two works. One is Vitpose\cite{xu2022vitpose} with a plain vision transformer, which achieves good performance. Another is BlazePose\cite{bazarevsky2020blazepose}, which is a lightweight architecture designed for mobile devices. In this paper, we directly use them as our Pose Estimation Network.
\section{Pose Saliency Annotation}

\begin{table*}[t]
\centering
\begin{tabular}{ c|c|c|c|c|c|c  }
\hline
\multirow{3}{*}{Action class} & \multicolumn{6}{|c}{RepCount-pose} \\
\cline{2-7}
& \multirow{2}{*}{Salient pose I} & \multirow{2}{*}{Salient pose II} & \multicolumn{2}{|c|}{training set} & \multicolumn{2}{|c}{test set} \\
\cline{4-7}
& & & video & event & video & event \\
\hline
bench press  & lying flat and arms-upward & lying flat and arms-down & 41 & 190 & 19 & 219 \\
front raise  & arms-down & arms-raise & 76 & 370 & 18 & 132 \\
jumping jack & body-upright and arms-down & jumping up and arms-upward & 49 & 350 & 26 & 713\\
pommel horse & body leaning to the left & body leaning to the right & 57 & 424 & 15 & 438 \\
sit-up       & lying down & sitting & 54 & 242 & 20 & 270 \\
squat        & body-upright & squatting & 81 & 544 & 18 & 164 \\
pull-up      & arms-hanging & arms-pull-up & 63 & 348 & 19 & 217 \\
push-up      & lying prostrate and arms-straight & lying prostrate and arms-bent & 66 & 449 & 16 & 303 \\
\hline
All & - & - & 487 & 2917 & 151 & 2456 \\
\hline
\end{tabular}
\vspace{1em}
\caption{Detailed information of \emph{RepCount-pose}, including salient poses, video count and event count of each action.}
\label{tab1}
\end{table*}

\begin{table*}[t]
\centering
\begin{tabular}{ c|c|c|c|c|c|c  }
\hline
\multirow{3}{*}{Action class} & \multicolumn{6}{|c}{UCFRep-pose} \\
\cline{2-7}
& \multirow{2}{*}{Salient pose I} & \multirow{2}{*}{Salient pose II} & \multicolumn{2}{|c|}{training set} & \multicolumn{2}{|c}{test set} \\
\cline{4-7}
& & & video & event & video & event \\
\hline
bench press & lying flat and arms-upward & lying flat and arms-down & 15 & 28 & 2 & 4 \\
body weight squat & body-upright & squatting & 19 & 50 & 4 & 9 \\
handstand push-up & handstanding and arms-straight & handstanding and arms-bent & 18 & 48 & 5 & 17  \\
jumping jack & body-upright and arms-down & jumping up and arms-upward & 17 & 49 & 5 & 20  \\
pommel horse & body leaning to the left & body leaning to the right & 20 & 66 & 5 & 48  \\
\hline
All & - & - & 89 & 241 & 21 & 98 \\
\hline
\end{tabular}
\vspace{1em}
\caption{Detailed information of \emph{UCFRep-pose}, including salient poses, video count and event count of each action.}
\label{tab2}
\end{table*}

There are some existing datasets for the repetition action counting task, such as \emph{UCFRep}\cite{zhang2020context} and \emph{Countix}\cite{dwibedi2020counting}. However, the biggest shortcoming is that only coarse-grained ground truth annotation are provided. Later, the proposal of \emph{RepCount}\cite{hu2022transrac} brings fine-grained annotation for the first time, in which start and end time of every action cycle are annotated to promote the further development in this field. It needs to be pointed out that such fine-grained annotation is very valuable when using video-level algorithms. 

% \begin{figure}[b]
% \centering
% \includegraphics[width=0.4\textwidth]{figure4.pdf}
% \caption{Action Distributions of {\bf \emph{RepCount-pose}} training set (left) and {\bf \emph{UCFRep-pose}} training set (right).}
% \label{fig4}
% \end{figure}

We introduce our pose-level model to learn a unique mapping between salient poses and actions. However, obtaining the core information of the frame indices where the salient poses occur is a challenge since such annotations are not available in current datasets. The most salient poses typically appear in the middle of the actions, so using only the start and end of actions in \emph{RepCount} may not allow the model to learn the mapping relationship effectively. Moreover, other datasets lack any fine-grained annotations, making it difficult to implement our pose-level method.

Based on the limitations of current datasets, we propose a novel {\bf Pose Saliency Annotation} that addresses the lack of annotations for salient poses. As Figure \ref{fig3} shows, we pre-define two salient poses for each action and annotate the frame indices where these poses occur for all videos in the training set, creating new annotation files for our pose-level method to train on. We apply this approach to two datasets, \emph{RepCount} and \emph{UCFRep}, and create two new annotated version called {\bf \emph{RepCount-pose}} and {\bf \emph{UCFRep-pose}}.

For {\bf \emph{RepCount-pose}}, after cleaning, the training set contains 487 videos and 2917 annotated action events. We retain the original test set without modification to test the robustness of our method. This new annotation scheme provides a foundation for pose-level methods and opens up new research opportunities. It can be observed that the number of events in the training set is close to the test set. This is because the pose-level method does not need to predict the number of repetitions during training, but only completes the mapping between salient poses and actions, so we do not need to capture every action event in the training set, but choose high-quality actions. In this regard, the cost of annotation will also be less than video-level methods. For {\bf \emph{UCFRep-pose}}, we selected 5 classes from the original 23 classes and annotated 110 videos to create a smaller dataset. Our pose-level model is lightweight enough to work effectively with such small dataset. The detailed information of these two new datasets are shown in Table \ref{tab1} and \ref{tab2}.

% Moreover, as Figure \ref{fig4} shows, the action distributions of two annotated datasets are relatively uniform, and we balanced the distribution without overcompressing the numbers of some higher frequency classes by selecting actions with higher-quality poses.

\section{PoseRAC Model}
\label{sec4}

\begin{figure*}[t]
\centering
\includegraphics[width=1.0\textwidth]{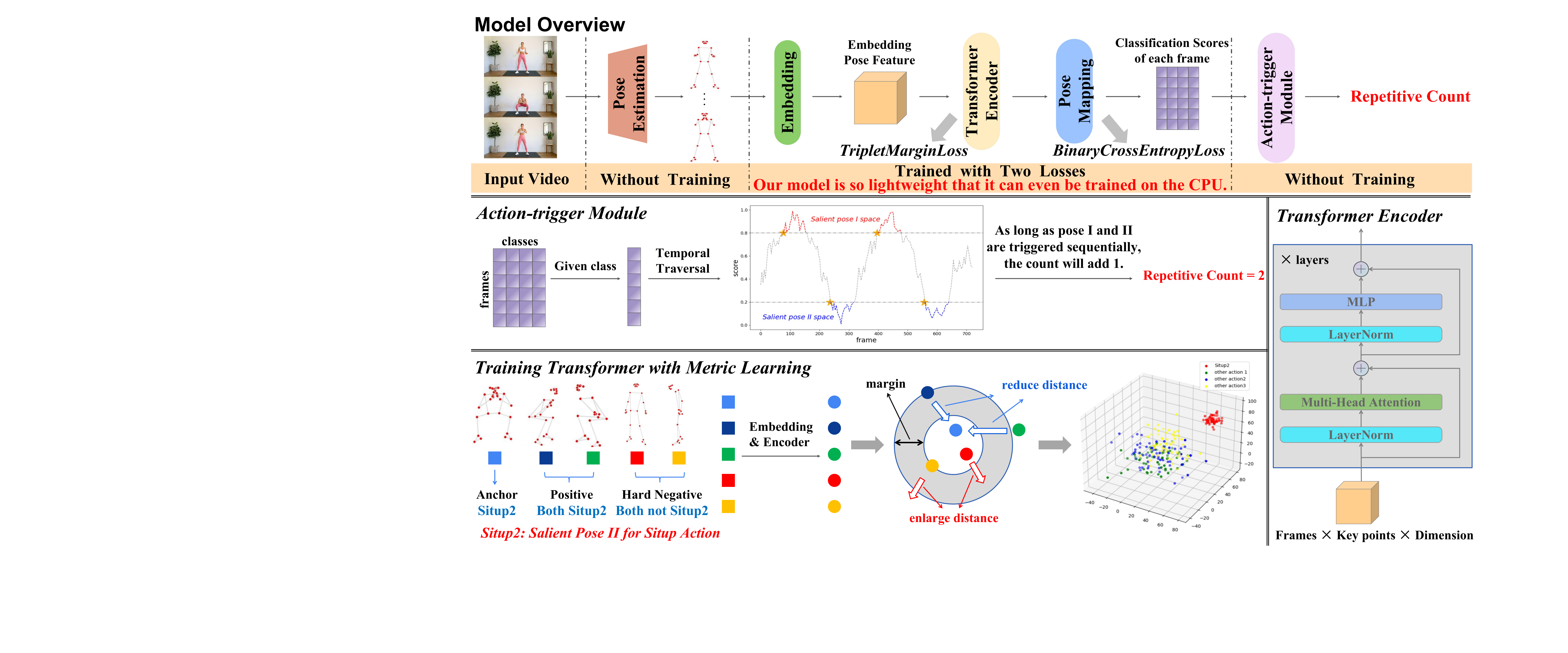}
\caption{Overview of our proposed PoseRAC. For a input video, the repetitive count can be obtained through Pose Estimation, Transformer Encoder, Pose Mapping and Action-trigger, where only the Encoder and the Pose Mapping need to be trained. We use Triplet Margin Loss to train the Encoder while Binary Cross Entropy Loss to train both the Encoder and the Pose Mapping. In addition to achieving the state-of-the-art performance so far, the biggest highlight of our PoseRAC is that it is lightweight enough to be easily trained on a CPU.}
\label{fig5}
\end{figure*}

Given a video $V={\{x_i\}}^{T}_{1}\in \mathbb{R}^{C\times H\times W\times T}$ with $T$ RGB frames, repetitive action counting model aims to predict a certain value $Y$, which is the number of repetitive actions. In this section, we will introduce our PoseRAC in detail.

\subsection{Model Overview}

As shown in Figure \ref{fig5}, PoseRAC consists of four parts. 

\begin{itemize}

\item The first is a state-of-the-art and lightweight Pose Estimation Network~($\S\ref{first}$), which is used to estimate the poses represented by lots of human pose key points from each frame of the original video sequence. 

\item The second is a simple Transformer Encoder~($\S\ref{second}$) to embed the key points of poses into high-level feature space, where the same class have similar distances, while the distances of different classes are far apart.

\item The third is a Pose Mapping Module~($\S\ref{third}$), where the unique mapping relationship between the salient poses and the action classes can be learned. Each pose can be mapped to the action class with the highest probability after the previous encoding.

\item The fourth part is a lightweight Action-trigger Module~($\S\ref{fourth}$). When we get the salient action classification results of all frames of the entire video sequence, we can use this module to calculate the repetition count in a short time.

\end{itemize}

\subsection{Pose Estimation Network}
\label{first}
Our model first converts the video sequence into a sequence of human pose key points, which can be defined as: 
\begin{equation}
\begin{split}
&V={\{x_i\}}^{T}_{1}\in \mathbb{R}^{C\times H\times W\times T}\\
&V\xrightarrow{\mathrm{Pose Estimation}} P={\{p_i\}}^{T}_{1}\in \mathbb{R}^{D\times K\times T}
\end{split}
\label{eq1}
\end{equation}
where each $x_i$ represents a single RGB frame, and each $p_i$ represents the key points of each frame. To express the key points of each frame, we use $D\times K$ sequence, which includes two parts, one ($K$) is the number of key points to fully represent the current pose, the other ($D$) is the dimension of each key point, generally three, which are the two coordinates of the planes and the depth estimation.

Here we use state-of-the-art pose estimation models such as Vitpose\cite{xu2022vitpose} and BlazePose\cite{bazarevsky2020blazepose}. The pose estimation algorithms themselves are not designed by us, but we introduce pose information into the action counting task, which is a novel design not explored by previous work.

Moreover, our pose-level poses estimation processes the primitive information of video, which is similar to the feature extraction network in all video-level algorithms such as I3D\cite{carreira2017quo}, VideoSwinTransformer\cite{liu2022video}, and TSN\cite{wang2016temporal}. But the difference is that the result of video-level incorporates all information, while pose-level only produces core information, which greatly improves the performance. Additionally, using pose information can contribute to the lightweight of model. For instance, for a 1024-frame video, video-level feature extraction with an output dimension of 512 would produce a data volume of $1024\times 512=524288$, while using pose information with 33 key points produces a data volume of only $1024\times 33 \times 3=101376$.

\subsection{Encoding Poses with Transformer}
\label{second}
Here we specify our data representation for the Transformer Encoder, which requires input batch size, sequence length, and embedding dimensions. In our pose-level approach, each frame is a batch, the number of key points in each frame is the sequence length, and the feature dimension of each key point is the embedding dimension.

First we get the pose of each frame ${p_i}\in \mathbb{R}^{D\times K}$ through the Pose Estimation Network, where $i\in {1, 2, \dots, T}$ is the frame index, $K$ is the number of key points, and $D$ is the dimension of each key point. We further define $p_i = {\{k_j\}}^{K}_{1}$ to represent each key point, where $k_j\in \mathbb{R}^D$, and we embed it to obtain richer information. Our embedding projection $\mathrm{\bf{E}}$ is a simple MLP network with ReLU as the activation function. These calculations can be defined as:
\begin{equation}
\begin{split}
\mathrm{\bf{Z}}^0 = [\mathrm{\bf{E}}(k_1), \mathrm{\bf{E}}(k_2), \dots, \mathrm{\bf{E}}(k_K)]^T
\end{split}
\end{equation}
where $\mathrm{\bf{E}}(k_j)\in \mathbb{R}^{D^{\prime}}$ is the embedding feature. Then the next Transformer takes $\mathrm{\bf{Z}}^0$ as input and encodes it with self-attention. Given $\mathrm{\bf{Z}}^0\in \mathbb{R}^{K\times D^{\prime}}$ with $K$ key point features, each of which is $D^{\prime}$-dimensional, $\mathrm{\bf{Z}}^0$ is projected using $\mathrm{\bf{W}}_Q\in \mathbb{R}^{D^{\prime}\times D_q}$, $\mathrm{\bf{W}}_K\in \mathbb{R}^{D^{\prime}\times D_k}$, $\mathrm{\bf{W}}_V\in \mathbb{R}^{D^{\prime}\times D_v}$, where $D_k=D_q$, to extract feature representations query($\mathrm{\bf{Q}}$), key($\mathrm{\bf{K}}$) and value($\mathrm{\bf{V}}$), which can be defined as:
\begin{equation}
\begin{split}
&\mathrm{\bf{Q}}=\mathrm{\bf{Z}}^0\times \mathrm{\bf{W}}_Q\\
&\mathrm{\bf{K}}=\mathrm{\bf{Z}}^0\times \mathrm{\bf{W}}_K\\
&\mathrm{\bf{V}}=\mathrm{\bf{Z}}^0\times \mathrm{\bf{W}}_V
\end{split}
\end{equation}
and the output of self-attention can be computed as:
\begin{equation}
\begin{split}
\mathrm{\bf{Attn}}=\mathrm{Softmax}(\frac{\mathrm{\bf{Q}}\mathrm{\bf{K}}^T}{\sqrt{D_q}})\mathrm{\bf{V}}
\end{split}
\end{equation}
where $\mathrm{\bf{Attn}}\in \mathbb{R}^{K\times D^{\prime}}$. Also, we use common multi-head self-attention (MHSA) to make several self-attention operations calculate in parallel.

Now we introduce the overall architecture of Transformer Encoder, which has $L$ layers with each layer consisting of MHSA and MLP blocks. Also, LayerNorm and Residual Connection are applied before and after every MHSA or MLP block, respectively. Because the number of key points of each frame is  a bit less, so our encoder does not include the downsampling module that other models may have. The overall process can be defined as:
\begin{equation}
\begin{split}
&\mathrm{\bf{\hat{Z}}}^l = \mathrm{MHSA}(\mathrm{LN}(\mathrm{\bf{Z}}^{l-1})) + \mathrm{\bf{Z}}^{l-1}\\
&\mathrm{\bf{Z}}^l = \mathrm{MLP}(\mathrm{LN}(\mathrm{\bf{\hat{Z}}}^l)) + \mathrm{\bf{\hat{Z}}}^l
\end{split}
\end{equation}
where $\mathrm{\bf{Z}}^{l-1}$, $\mathrm{\bf{\hat{Z}}}^l$, $\mathrm{\bf{Z}}^l\in \mathbb{R}^{K\times D^{\prime}}$.

\subsection{Pose Mapping}
\label{third}
Taking the Encoder output $\mathrm{\bf{Z}}^L\in \mathbb{R}^{K\times D^{\prime}}$ as input, Pose Mapping module outputs probability scores $\mathrm{\bf{S}}\in \mathbb{R}^{C}$ of the current frame over all action classes. We perform binary classification after Sigmoid activation for each class, with the two salient poses of each class represented by the same bit data. To realize such a module, we use a very lightweight MLP network, which avoids the complexity. First, the two dimensions $K$ and $D^{\prime}$ of $\mathrm{\bf{Z}}^L$ are flattened into $\mathbb{R}^{KD^{\prime}}$, and then it passes through an MLP module, where the output channels is set to $C$, which can be defined as:
\begin{equation}
\begin{split}
\mathrm{\bf{S}} = \sigma(\mathrm{MLP}(\mathrm{Flatten}(\mathrm{\bf{Z}}^L)))
\end{split}
\end{equation}
where $\sigma$ represents the Sigmoid activation function.

With such Pose Mapping, we can obtain the scores of single frame. It should be noted that we extract the poses of all frames, and use the convenience of matrix operations to obtain scores in parallel, which is actually consistent with the idea of mini batch. So at last, we combine the scores of all frames to get the video score matrix $\mathrm{\bf{\hat{S}}}\in \mathbb{R}^{C\times T}$, where $T$ represents the number of frames in the current video.

\subsection{Action-trigger Module}
\label{fourth}
We use the lightweight Action-trigger Module to obtain the final output $Y$, the repetitive action count, which has a time complexity of $\mathcal{O}(n)$. First, we get the scores $S_c\in \mathbb{R}^T$ of a given action class from $\mathrm{\bf{\hat{S}}}$. Then, we scan all frames and use the action-trigger mechanism to count when the two salient poses of the action class occur sequentially. We set upper and lower bounds to distinguish the scores of the two salient poses, which cluster non-salient poses in the middle and easily classify the salient poses to the two ends.

\subsection{Losses and Metric Learning}

The modules need to be trained are Embedding, Transformer Encoder and Pose Mapping, and because we perform binary classification for each class, so we use the Binary Cross Entropy Loss, which can be defined as follows:
\begin{gather}
\mathcal{L}_{bce} = -\frac{1}{N}\sum\limits_{i=1}^{N}(\frac{1}{C}\sum\limits_{j=1}^{C}loss(i,j))  \\
 loss(i,j)=y_{ij}\log p_{ij} + (1-y_{ij})\log(1-p_{ij})
\end{gather}
where $N$ represents the batch size (in our method, each frame is a batch), $C$ represents the number of classes, $y$ and $p$ are the labels and our predictions, respectively.

Moreover, we use Metric Learning to improve our Encoder and introduce the Pose Triplet Loss. Given a pose, Encoder produces higher-level features $\mathrm{\bf{Z}}^L$, which should be more representative. As shown in Figure \ref{fig5}, we achieve this with Triplet Margin Loss function, which selects anchors, same class positive samples, and different classes negative samples in a batch. It can be expressed as:
\begin{equation}
\begin{split}
\mathcal{L}_{tri} = \mathrm{max}(\mathrm{CS}(a,p)-\mathrm{CS}(a,n)+\mathrm{margin},0)
\end{split}
\end{equation}
where $a$, $p$, $d$ are anchors, positive and negative samples, and $\mathrm{CS}$ represents the Cosine Similarity to measure the distance between features. We pay more attention to hard samples, where the distances between anchors and negative samples are even smaller than those of positive samples. After Metric Learning, the poses of each action can be distinguishable, which cluster in the high-level space.

At last, our overall training combines these two losses:
\begin{equation}
\begin{split}
\mathcal{L} = \mathcal{L}_{bce} + \alpha\mathcal{L}_{tri}
\end{split}
\end{equation}
where $\alpha$ is the weight factor to control the two losses in the same numeric scale.
\subsection{Implementation Details}

\noindent{\bf Training.} We use the \emph{RepCount-pose} and \emph{UCFRep-pose} dataset we created to train our model. Only the frames with salient poses are inputted into the network instead of the entire video to speed up the fitting.

\noindent{\bf Inference.} During inference, the entire video sequence is inputted into the model. The poses of all frames pass through the Encoder and Pose Mapping, and then enter the Action-trigger Module to output the repetitive count.
\section{Experiments and Results}

\begin{table*}[t]
\centering
\begin{tabular}{ c |c||c|c||c|c||c }
\hline
\multicolumn{2}{c||}{\multirow{2}{*}{Methods}} & \multicolumn{2}{c||}{RepCount (-pose)} & \multicolumn{2}{c||}{UCFRep (-pose)} & \multirow{2}{*}{Time (ms)} \\
\cline{3-6}
\multicolumn{2}{c||}{} & MAE $\downarrow$ & OBO $\uparrow$  & MAE $\downarrow$ & OBO $\uparrow$ & \\
\hline
\multirow{7}{*}{video-level} & RepNet\cite{dwibedi2020counting} & 0.995 & 0.013 & 0.981 & 0.018 & 100\\
& X3D\cite{feichtenhofer2020x3d} & 0.911 & 0.106 & 0.982 & 0.331 & 220\\
& Zhang et al.\cite{zhang2020context} & 0.879 & 0.155 & 0.762 & 0.412 & 225\\
& TANet\cite{liu2021tam} & 0.662 & 0.099 & 0.892 & 0.129 & 187\\
& VideoSwinTransformer\cite{liu2022video} & 0.576 & 0.132  & 1.122 & 0.033 & 149\\
& Huang et al.\cite{huang2020improving}  & 0.527 & 0.159  & 1.035 & 0.015 & 156\\
& TransRAC\cite{hu2022transrac}  & 0.443 & 0.291 &  0.581 & 0.329 & 200\\
\hline
first pose-level & {\bf PoseRAC (Ours)} & {\bf 0.236} & {\bf 0.560}  & {\bf 0.312}  & {\bf 0.452} & {\bf 20}\\
\hline
\end{tabular}
\vspace{1em}
\caption{Performance on \emph{RepCount(-pose)} and \emph{UCFRep(-pose)} test. For the \emph{RepCount} and \emph{RepCount-pose} (also for the \emph{UCFRep} and \emph{UCFRep-pose}), their test sets are of the same, and apart from the difference in annotations, their training sets are also the same, as our pose-level method requires pose saliency annotation, while all other video-level methods require traditional boundary annotation.}
\label{tab3}
\end{table*}

We present experiments and results on \emph{RepCount} and \emph{UCFRep} benchmarks, which we upgrade to \emph{RepCount-pose} and \emph{UCFRep-pose} using our proposed Pose Saliency Annotation. This new design enables our proposed pose-level methods to provide a fair comparison with previous state-of-the-art methods. The main evaluation metrics used in previous work\cite{dwibedi2020counting, zhang2020context, hu2022transrac} are \textbf{Off-By-One (OBO) count error} and \textbf{Mean Absolute Error (MAE)}. OBO measures the error rate of repetition count over the entire dataset, while MAE represents the normalized absolute error between the ground truth and the prediction. They can be defined as:
\begin{gather}
\mathrm{\bf{OBO}} = \frac{1}{N}\sum\limits_{i=1}^N[\vert \Tilde{c_i}-c_i\vert\leq 1]\\
\mathrm{\bf{MAE}} = \frac{1}{N}\sum\limits_{i=1}^N\frac{\vert \Tilde{c_i}-c_i\vert}{\Tilde{c_i}}
\end{gather}
where $\Tilde{c}$ is the ground truth, $c_i$ is our prediction, and $N$ is the number of videos.

\subsection{Experiment Setup}
Our hardware setup includes an Intel Core i7 Xeon CPU, GeForce RTX 3090 Ti GPU, and 64 GB RAM. We use PyTorch to implement our method, which utilizes the lightweight BlazePose\cite{bazarevsky2020blazepose} for Pose Estimation. Our model has a simple fully-connected layer as Embedding, a six-layer Transformer Encoder, and a two-layer MLP network as Pose Mapping. This lightweight model trains quickly, with 15 epochs completed in 20 minutes on a GPU or one hour and a half on a CPU.

\subsection{Evaluation and Comparison}

Our approach for repetitive action counting, PoseRAC, outperforms existing methods in terms of both accuracy and speed. Table \ref{tab3} shows that our method consistently outperforms previous methods on both datasets and under both evaluation metrics, with an OBO metric of 0.56 compared to the 0.29 of TransRAC under the \emph{RepCount-pose}. Additionally, our lightweight input data and model make our method significantly faster than previous video-level methods. In the inference stage, our method has the fastest average processing time per frame, with nearly ten times the speed improvement over TransRAC, such superiority of our PoseRAC can be seen in Figure \ref{fig6}.

\subsection{Ablation Studies}

We conduct ablation studies on \emph{RepCount-pose} to analyze three core ideas of PoseRAC.

\noindent{\bf Choice of Pose Estimation Network.} The first part of PoseRAC is a Pose Estimation Network, and we compare two excellent algorithms: Vitpose\cite{xu2022vitpose} and BlazePose\cite{bazarevsky2020blazepose} in Table \ref{tab4}. Although Vitpose has more powerful learning ability, BlazePose estimates an additional depth information, resulting in better performance when applied to this task. Moreover, BlazePose is designed for mobile terminals, offering an advantage in speed. Therefore, we choose BlazePose as the Pose Estimation Network for our work.

\begin{figure}[b]
\centering
\includegraphics[width=0.4\textwidth]{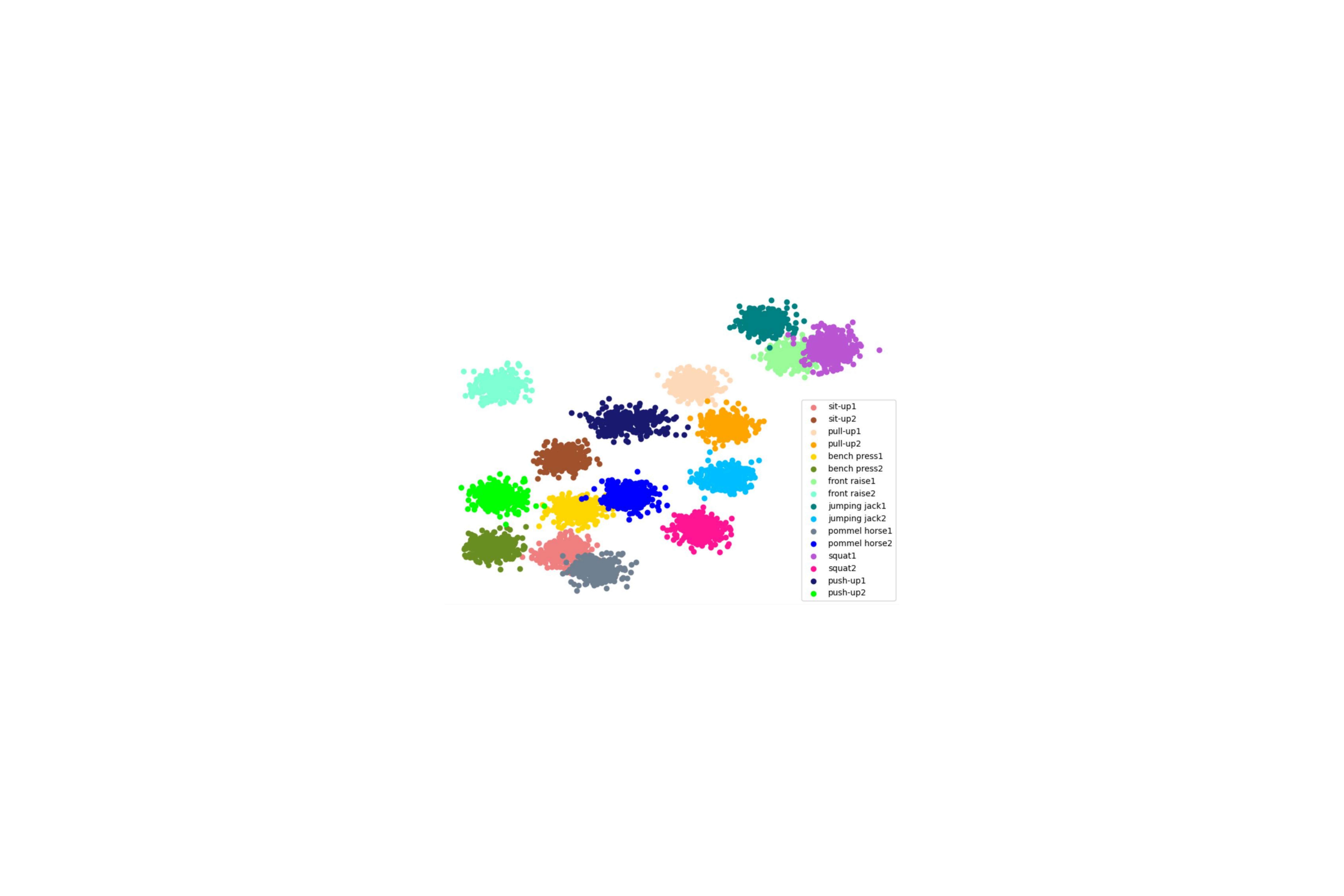}
\caption{The distribution of embedding feature for each action in high-level space after training with Metric Learning.}
\label{fig7}
\end{figure}

\begin{table}[t]
\centering
\begin{tabular}{ c|c|c|c }
\hline
Algorithms & MAE $\downarrow$ & OBO $\uparrow$  & Time (ms) \\
\hline
BlazePose\cite{bazarevsky2020blazepose} & {\bf 0.236} & {\bf 0.560} & {\bf 20} \\
Vitpose\cite{xu2022vitpose}   & 0.305 & 0.463 & 46 \\
\hline
\end{tabular}
\vspace{1em}
\caption{Comparison of different pose estimation algorithms.}
\label{tab4}
\end{table}

\begin{table}[t]
\centering
\begin{tabular}{ c|c|c|c }
\hline
Loss & $\alpha$ & MAE $\downarrow$ & OBO $\uparrow$ \\
\hline
$\mathrm{\bf{L}}_{cls}$ only & - & 0.317 & 0.486 \\
\hline
\multirow{3}{*}{$\mathrm{\bf{L}}_{cls} + \alpha\mathrm{\bf{L}}_{tri}$} & {\bf 0.01} & {\bf 0.236} & {\bf 0.560}  \\
& 0.05 & 0.289 & 0.501 \\
& 0.1 & 0.328 &  0.462 \\
\hline
\end{tabular}
\vspace{1em}
\caption{The effect of Metric Learning.}
\label{tab5}
\end{table}

\begin{table}[t]
\centering
\begin{tabular}{ c|c|c }
\hline
Methods & MAE $\downarrow$ & OBO $\uparrow$ \\
\hline
ResNet-50\cite{he2016deep} & 0.556 & 0.181 \\
ViT-32\cite{dosovitskiy2020image} & 0.517 & 0.195 \\
\hline
{\bf PoseRAC} & {\bf 0.236} & {\bf 0.560} \\
\hline
\end{tabular}
\vspace{1em}
\caption{Comparison of common image classification methods and our pose-level method.}
\label{tab6}
\end{table}

\noindent{\bf Effectiveness of Metric Learning.} We use two losses to train our model. Table \ref{tab5} compares the performance with and without Triplet Margin Loss using different values of $\alpha$. Our model can be effectively trained with Binary Cross Entropy Loss only, but adding Metric Learning improves it, and the best value of $\alpha$ is found to be 0.01. Our Metric Learning improves the optimization, and Figure \ref{fig7} shows that the Encoder trained with Metric Learning enhances the ability to distinguish salient poses of each class.

\noindent{\bf Image classification and our pose-level method.} 
Our pose-level method outperforms those based on image classification. Simply replacing our Encoder and Pose Mapping with an image classification algorithm would result in a severe drop in performance, as shown in Table \ref{tab6}. While our method extracts the core information, i.e., the pose of each frame, image classification methods bring in irrelevant information, which is similar to the video-level methods.

\begin{figure}[t]
\centering
\includegraphics[width=0.47\textwidth]{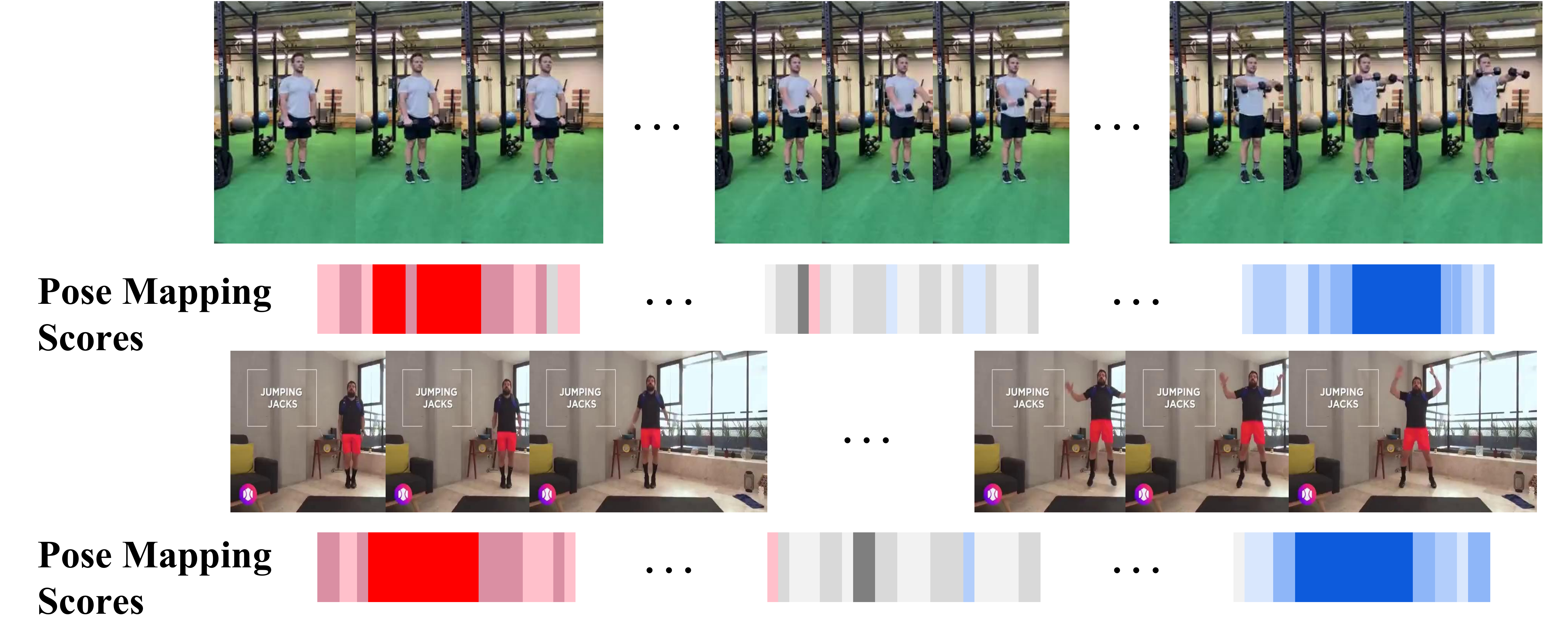}
\caption{Visualization of Pose Mapping. Under a given class, for each frame, the higher the score (the darker the red in figure), the more likely it is salient pose I, and the lower the score (the darker the blue), the more likely it is salient pose II. When the score is closer to the middle (the gray part in the figure), the pose of current frame is more likely to be an irrelevant pose. We can see that the prediction results of the model are very accurate, as when the salient posture I or II occurs in the video, the color is red or blue, respectively.}
\label{fig8}
\end{figure}

\subsection{Qualitatively Evaluation}
To verify the effectiveness of our method, we visually analyze the output of Pose Mapping. As shown in Figure \ref{fig8}, trained with Pose Saliency Annotation, PoseRAC has a strong ability to discriminate the salient poses of various action classes, that is, when these poses occur in the video, our model can accurately recognize them and determine the corresponding space. Specifically, the score of salient pose I is greater than 0.8 (red part in the figure), the score of salient pose II is less than 0.2 (blue part in the figure), and other irrelevant poses have an output close to 0.5 (gray part in the figure).
\section{Conclusion}
In conclusion, this paper presents a significant contribution to the field of repetitive action counting by introducing the novel approach, Pose Saliency Representation, which efficiently represents each action using only two salient poses. The proposed pose-level method, PoseRAC, based on this representation, achieves state-of-the-art performance on two new version datasets by utilizing Pose Saliency Annotation for training. Our lightweight model requires only 20 minutes for training on a GPU and infers nearly 10x faster compared to previous methods, making it highly efficient for practical use. Moreover, our approach significantly outperforms the previous state-of-the-art TransRAC, achieving an OBO metric of 0.56 compared to the 0.29 of TransRAC, demonstrating the effectiveness of our proposed method. The code and new version dataset are publicly available, enabling the research community to reproduce our results and conduct further experiments. Overall, our approach shows promising results and opens up new avenues for future research in the field of repetitive action counting.

{\small
\bibliographystyle{ieee_fullname}
\bibliography{egbib}
}

\end{document}